\documentclass{article}

\PassOptionsToPackage{numbers, compress}{natbib}

\usepackage[preprint]{neurips_2025}

\usepackage[utf8]{inputenc} 
\usepackage[T1]{fontenc}    
\usepackage{hyperref}       
\usepackage{url}            
\usepackage{booktabs}       
\usepackage{amsfonts}       
\usepackage{nicefrac}       
\usepackage{microtype}      
\usepackage{xcolor}         
\usepackage{todonotes}
\usepackage{amsmath}
\usepackage{algorithm}
\usepackage{multirow}
\usepackage{algpseudocode}

\algtext*{EndFor}

\hypersetup{
  colorlinks=True,
  citecolor=green,
  linkcolor=red,
  urlcolor=blue
}

\title{Towards Understanding Best Practices for Quantization of Vision-Language Models}

\author{%
  Gautom Das$^{*}$ \quad Vincent La$^{*}$ \quad Ethan Lau$^{*}$ \quad Abhinav Shrivastava \quad Matthew Gwilliam \\
  University of Maryland, College Park \\
}

\begin{document}

\def\thefootnote{*}\footnotetext{Equal contribution.}

\maketitle

\begin{abstract}
    Large language models (LLMs) deliver impressive results for a variety of tasks, but state-of-the-art systems require fast GPUs with large amounts of memory.
    To reduce both the memory and latency of these systems, practitioners quantize their learned parameters, typically at half precision.
    A growing body of research focuses on preserving the model performance with more aggressive bit widths, and some work has been done to apply these strategies to other models, like vision transformers.
    In our study we investigate how a variety of quantization methods, including state-of-the-art GPTQ and AWQ, can be applied effectively to multimodal pipelines comprised of vision models, language models, and their connectors.
    We address how performance on captioning, retrieval, and question answering can be affected by bit width, quantization method, and which portion of the pipeline the quantization is used for.
    Results reveal that ViT and LLM exhibit comparable importance in model performance, despite significant differences in parameter size, and that lower-bit quantization of the LLM achieves high accuracy at reduced bits per weight (bpw). 
    These findings provide practical insights for efficient deployment of MLLMs and highlight the value of exploration for understanding component sensitivities in multimodal models.
    Our code is available at https://github.com/gautomdas/mmq.
\end{abstract}

\section{Introduction}

As neural networks grow larger, especially with the rise of large language models (LLMs)~\cite{kaplan2020scalinglawsneurallanguage}, it has become increasingly important to reduce memory and computational demands~\cite{nagel2021white}. 
Training these models demands massive GPU/TPU clusters, and even inference can be costly~\cite{naveed2024comprehensiveoverviewlargelanguage}. 
Additionally, distributed multi-GPU setups are not practical for model deployment on edge devices and in similar real-world scenarios~\cite{dhar2024empirical}. 
As these models become more ubiquitous, their cost and accessibility become massively important societal issues.

While text-only LLMs are expensive enough on their own, real-world applications increasingly require models that can process not only text, but also audio and visual data, leading to the rise of multimodal LLMs (MLLMs)~\cite{li2023blip2bootstrappinglanguageimagepretraining,liu2023visualinstructiontuning}.
Multimodal large language models (MLLMs), which combine LLMs with vision encoders (and/or audio encoders~\cite{shu2023audio}), require even more resources, and exacerbate cost and accessibility issues.
While these can deliver impressive performance on multimodal retrieval, captioning, and question-answering tasks, the performance typically scales with model size and latency.
Video data drives additional massive increases in latency and memory consumption.

Model compression, particularly quantization, offers a potential solution to the growing model size problem~\cite{nagel2021white}.
One goal of model compression is to find where ``free lunch'' ends -- how small we can make a model without sacrificing performance.
Another goal is to aggressively compress the model beyond this point, while mitigating the penalty to performance.
Quantization tackles this by preserving all model weights and their relative approximate values, but changing the precision at which these values are stored.

\begin{figure}[t]
	\centering
	\includegraphics[width=1.0\linewidth]{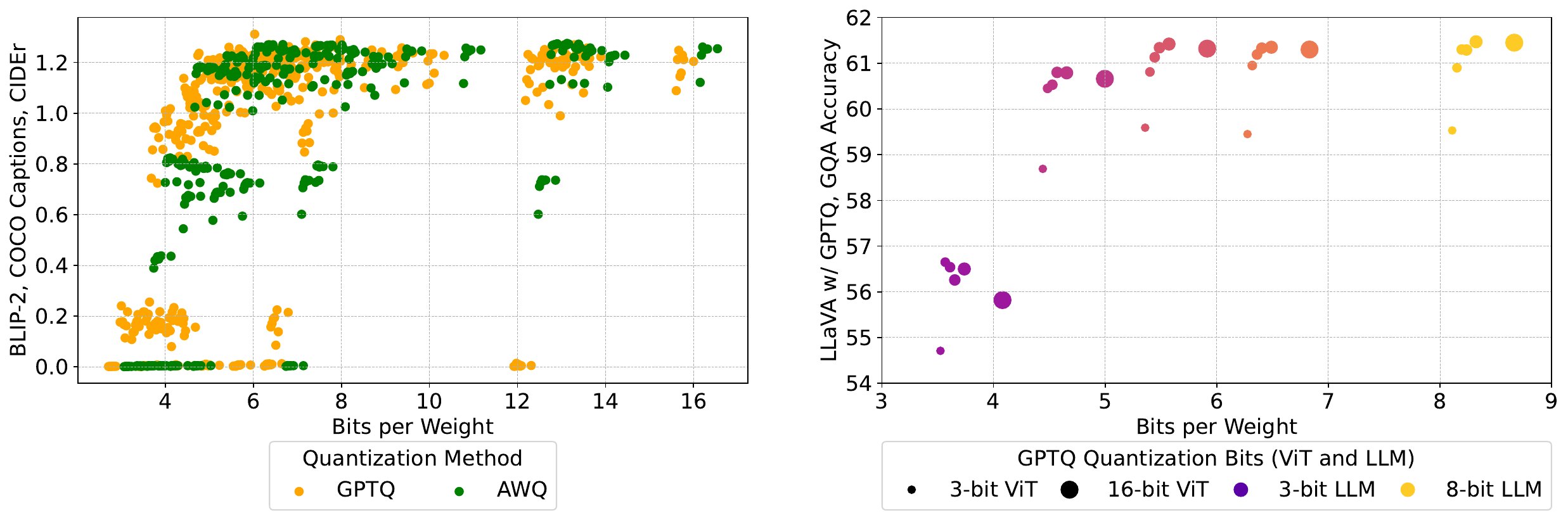}
	\caption{\textbf{Highlighted analysis.} In this work, we investigate how quantizations affects multimodal models on a variety of tasks, specifically BLIP-2 (left) and LLaVA (right). We compare state-of-the-art quantization strategies (left) and also discuss which portions of the pipeline are most amenable to quantization (right).}
	\label{fig:teaser}
\end{figure}

One might assume that if we uniformly quantize weights, model performance will decrease with some linear correlation to the reduction in bit width.
In reality, as we show in Figure~\ref{fig:teaser}, performance varies dramatically depending on which parts of an MLLM are quantized, and which method is chosen for the quantization.
Decisions like which method to use (as we compare state-of-the-arts GPTQ~\cite{frantar2023gptqaccurateposttrainingquantization} and AWQ~\cite{lin2024awqactivationawareweightquantization}) can dramatically affect performance at lower bit widths.
Quantizing the ViT compared to the LLM has dramatically different effects on the performance-size trade-off, considering that while both affect the performance, quantizing the LLM has a much higher impact on the size of the overall MLLM.
Understanding multimodal quantization is key to improving model efficiency, as it reveals which components are most critical for performance. 
Not all components are equally sensitive to a reduction in precision, and thus some are more tolerable to quantization techniques.
By minimizing information loss across salient model components, we can deploy quantized multimodal models that optimize the model size/task performance trade-off.

In summary, we distill several key principles governing  MLLMs and quantization strategies. First, component sensitivity varies substantially across multimodal architectures, with language models generally requiring higher precision than vision components regardless of their parameter count. Second, state-of-the-art methods like GPTQ and AWQ effectively preserve model performance at significantly lower bit widths (3.5-4.5 bpw) compared to uniform approaches. Third, task characteristics fundamentally determine optimal bit allocations—reasoning tasks heavily favor LLM precision while visual-textual alignment tasks show more balanced requirements. Fourth, the choice of quantization method dramatically redistributes component importance, with AWQ concentrating on LLM preservation while GPTQ distributes importance more evenly. Finally, architectural dependencies create interaction effects that necessitate holistic pipeline analysis rather than independent component evaluation. These principles provide practical guidance for efficient MLLM deployment across diverse tasks and computational constraints.

\section{Related Work}

\subsection{Vision Language Models}

Vision language models (VLMs) are a subset of multimodal models whose inputs and outputs comprise images, text, and sometimes videos.
These approaches are typically trained using video/image-text pairs to align vision and language inputs in a shared space~\cite{miech2019howto100m,radford2021learning,jia2021scaling,bain2022frozen,zhang2023visionlanguage}.
In this work, we make a distinction between these VLMs, which typically focus on generating embeddings of images and text for tasks like retrieval, and vision large language models (VLLMs), which typically consume image and text inputs to generate language outputs~\cite{li2023blip2bootstrappinglanguageimagepretraining,neurips_flamingo,neurips_frozen,mokady2021clipcapclipprefiximage,chen2023minigptv2,zhu2023minigpt,awadalla2023openflamingo}, such as answers to complicated questions about specific images~\cite{hudson2019gqanewdatasetrealworld,goyal2017makingvvqamatter}.
BLIP-2~\cite{li2023blip2bootstrappinglanguageimagepretraining} alleviates the expensive potential training cost of VLLMs by introducing a Q-Former to bridge the gap between frozen image models~\cite{radford2021learning,dosovitskiy2020image} and frozen language models~\cite{zhang2022optopenpretrainedtransformer}.
LLaVA~\cite{liu2023visualinstructiontuning} uses a simpler projector to connect the frozen models, but finetunes both using instruction tuning.
In this work, we investigate how quantization affects both BLIP-2 and LLaVA as representative VLLMs.

\begin{figure}[t]
    \begin{minipage}{0.55\linewidth}
	\centering
	\includegraphics[width=1.0\linewidth]{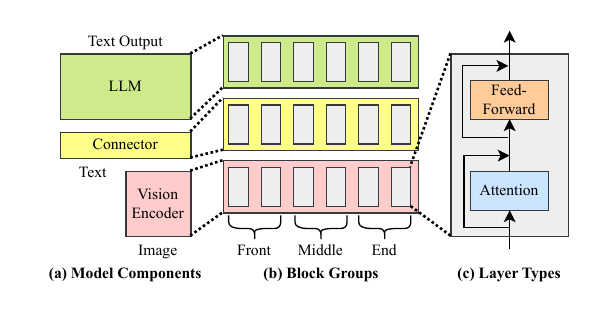}
    \end{minipage}
    \hfill
    \begin{minipage}{0.43\linewidth}
        \begin{algorithm}[H]
        \small
        \caption{Uniform Quantization Experiments}
        \begin{algorithmic}[1]
        \For{$k \in \{2,4,6,8\}$}
            \ForAll{$C \subseteq \{\text{ViT}, \text{LLM}, \text{QFormer}\}$}
                \ForAll{$B  \subseteq \{\text{front}, \text{middle}, \text{end}\}$}
                    \ForAll{$M  \subseteq \{\text{attn}, \text{FF}\}$}
                        \State $\text{quantize}(k, C, B, M)$
                    \EndFor
                \EndFor
            \EndFor
        \EndFor
        
        \end{algorithmic}
        \end{algorithm}
    \end{minipage}
	\caption{We investigate the sensitivity of the various parts of an MLLM to quantization, specifically in terms of model components, block groups, and layer types. For uniform quantization, we perform a dense grid search of bit widths and combinations of these components. For state-of-the-art methods (GPTQ and AWQ) we perform a more targeted analysis.}
	\label{fig:diagram}
\end{figure}

\subsection{Model Compression}

\noindent\textbf{Pruning} seeks to remove neurons of low saliency, often resulting in sparser computation. This can be done in an unstructured manner~\cite{NIPS1989_6c9882bb,dong2017learningprunedeepneural,park2020lookaheadfarsightedalternativemagnitudebased,NEURIPS2019_4efc9e02}, where individual weights are removed, wherever they may occur, or in a structured manner~\cite{He_2018,yu2018nisppruningnetworksusing,luo2017thinetfilterlevelpruning}, where entire layers are removed. While this can greatly reduce model size and FLOPs, it often comes at the cost of model accuracy. The pruning process can be done iteratively to monitor performance degradation, or fine-tuning can be done afterward to help recover the performance of the full-sized model.

\noindent\textbf{Knowledge distillation} (KD) involves training a smaller, compact student model to mimic the behavior of a larger teacher model. The student model uses logits from the larger teacher as soft targets during training~\cite{hinton2015distillingknowledgeneuralnetwork, ahn2019, li2017learningnoisylabelsdistillation, Yim_2017_CVPR, yin2020dreamingdistilldatafreeknowledge, gu2024minillmknowledgedistillationlarge}. KD often struggles to achieve high compression ratios on its own, when compared to pruning or quantization, as an overly compressed student model can struggle to approximate the complexity of the teacher model.

\noindent\textbf{Quantization} refers to a reduction in the numerical precision of neural network weights and activations~\cite{guo2018survey,hubara2018quantized}. 
It is often utilized for the deployment of models on edge devices, achieving more efficient inference by leveraging low-bit integer arithmetic~\cite{nagel2021white}.
Quantization methods tend to fall into two categories: post-training quantization (PTQ) and quantization-aware training (QAT).
PTQ involves quantizing weights/activations after training a full-precision model, often using a representative calibration set to determine scales, clipping ranges, and the saliency of parameters~\cite{frantar2023gptqaccurateposttrainingquantization,lin2024awqactivationawareweightquantization,nagel2020downadaptiveroundingposttraining,hubara2020improvingposttrainingneural,wang2020towards,li2021brecqpushinglimitposttraining,frantar2023optimalbraincompressionframework, dettmers2023spqrsparsequantizedrepresentationnearlossless, lee2024owqoutlierawareweightquantization, dettmers2022llmint88bitmatrixmultiplication}.
QAT involves quantizing a model and then training/fine-tuning a model to adjust parameters to recover model performance degradation~\cite{jacob2017quantizationtrainingneuralnetworks,liu2023llmqatdatafreequantizationaware, pmlr-v162-nagel22a, chen2024efficientqatefficientquantizationawaretraining, shao2024omniquantomnidirectionallycalibratedquantization}. 
PTQ is more lightweight than QAT as it does not require additional training.
Recent interest in extreme quantization of LLMs has even resulted in a third paradigm, related to QAT, where the architecture itself is designed with quantization in mind, utilizing ternary and even binary weights~\cite{wang2023bitnetscaling1bittransformers,ma2024era1bitllmslarge, xu2024onebitextremelylowbitlarge}.
Quantization can be combined with pruning or distillation of knowledge for even larger amounts of model compression~\cite{gholami2021surveyquantizationmethodsefficient,chen2024ternaryllmternarizedlargelanguage, du2024bitdistillerunleashingpotentialsub4bit}.  
In this work, we focus specifically on quantization methods which may require calibration, but do not require any additional training/fine-tuning.
These include uniform quantization, as well as state-of-the-art (SOTA) methods, GPTQ~\cite{frantar2023gptqaccurateposttrainingquantization} and AWQ~\cite{lin2024awqactivationawareweightquantization}.
Instead of applying these to unimodal models (LLMs), we investigate how well they work in the multimodal setting.

\begin{figure}[t]
	\centering
	\includegraphics[width=\linewidth]{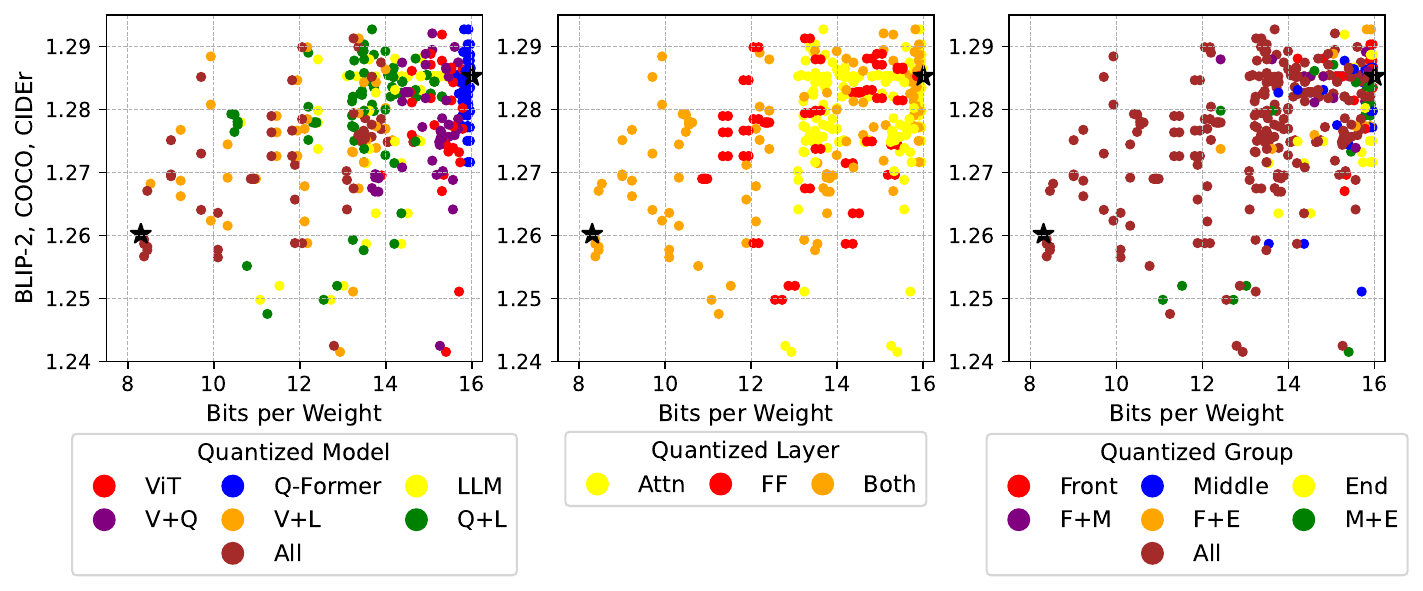}
	\caption{\textbf{Uniform quantization} impact on COCO captioning for BLIP-2.
    We focus these plots on the high-quality range (see appendix for full results), drawing attention first to which model component we quantize (left), then layer type (middle), then group (right).
    We provide 2 black stars on each plot, indicating the results for quantizing the entire pipeline at 8 and 16 bits.
    The best performance-size tradeoffs tend to come when quantizing the entire pipeline.
    }
	\label{fig:blip_uniform_trio}
\end{figure}

\section{Optimizing Performance}
\label{sec:optimizing_performance}

\subsection{Building Intuitions}
\label{subsec:building_intuitions}

For our preliminary experiment, we explore the impact of reduced bit widths on different parts of the BLIP-2 architecture via simulated uniform quantization.
At a high level, this procedure involves normalizing the weights, discretizing them, and then returning them to their original scale for inference. For the derivation of the $k$-bit uniform quantization formula used, please see the Appendix.

After selecting a bit width, $k$, from \{2,4,6,8\}, we subdivide the BLIP-2 architecture along different axes shown in Figure~\ref{fig:diagram}, and we investigate the impact of $k$-bit uniform quantization. 
First we consider ``model components'', which are the ViT and the Q-Former for retrieval, and the ViT, LLM, and Q-Former for captioning. We also can divide these in terms of ``block groups,'' where we split each model component into 3 equal-size, contiguous groups of blocks, which we call front, middle, and end.
Finally, we target the quantization by different ``layer types,'' applying it to varying combinations of attention (attn) and feedforward (FF) layers.

We show results for this exploration in Figure~\ref{fig:blip_uniform_trio}.
Overall, we do not observe a significant correlation between size and performance for layer types or block groups.
Simply quantizing both layer types and all block groups is necessary for best results.
Also, while simply quantizing the entire pipeline at 8 bits is a point along the Pareto-frontier, there are still many points along the frontier that do not quantize everything, especially in terms of model components.
These findings are largely consistent on the LLM-free retrieval task as well, as we show in the appendix.
Therefore, we focus the majority of our analysis in this section on the model components (ViT, Q-Former, LLM, or equivalents).

\begin{figure}[t]
	\centering
	\includegraphics[width=1.0\linewidth]{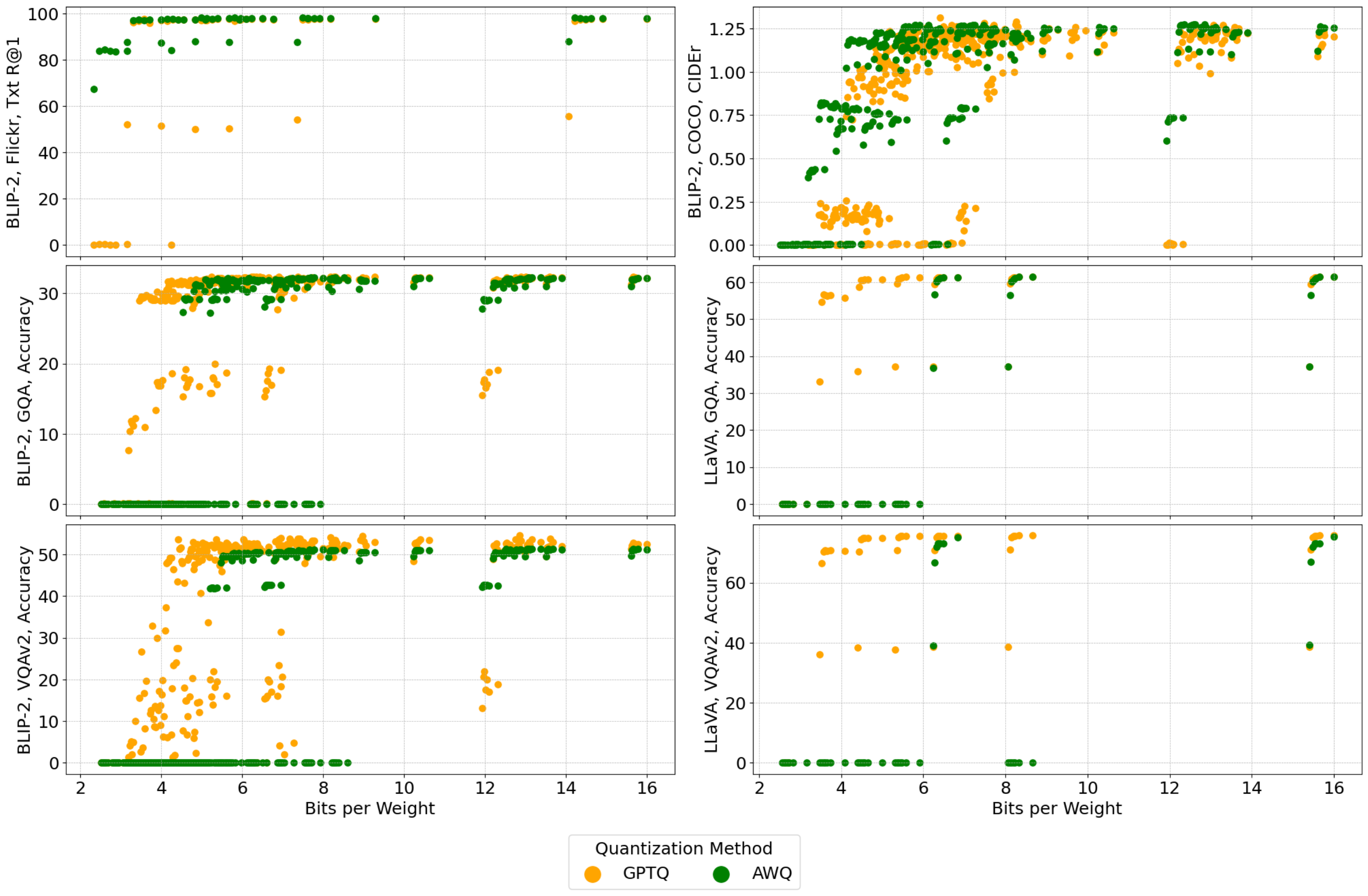}
	\caption{\textbf{SOTA Unimodal Quantization Performance-Size Tradeoff} for retrieval, captioning, and VQA tasks. We apply GPTQ and AWQ to entire model components (vision encoder, connector, LLM), of BLIP-2 and LLaVA. Unimodal SOTA methods are able to preserve full-precision performance at high compression rates, even in the multimodal context.}

\label{fig:sota_plots}
\end{figure}

\subsection{State-of-the-Art Quantization Preliminaries}
\label{subsec:sota_quant_prelim}

Following our preliminary quantization experiments, we apply state-of-the-art (SOTA) unimodal quantization methods, Activation-aware Weight Quantization (AWQ) and GPTQ to BLIP-2 and LLaVA.

\textbf{AWQ}~\cite{lin2024awqactivationawareweightquantization} is a weight-only PTQ technique for LLMs that identifies salient weight channels by referring to the activation distribution of a small, representative calibration set. Weights that yield greater activation magnitudes correspond to more critical features and are deemed to be more salient. AWQ reduces quantization error by preserving just 1\% of salient weights via a per-channel scaling factor.

\textbf{GPTQ}~\cite{frantar2023gptqaccurateposttrainingquantization} is another weight-only PTQ technique for LLMs but leverages approximate second-order information derived from the inverse Hessian instead of activation information. As weights are processed and quantized sequentially, the remaining unquantized weights are adjusted to compensate for quantization error, computed with a small calibration set. GPTQ replaces iterative Hessian updates with a Cholesky-based reformulation for greater numerical stability when handling billion-parameter LLMs.

Since we observe in Section~\ref{subsec:building_intuitions} that the choice of block group has little to no impact on quantized model performance, we conduct these experiments with a coarser granularity. 
We quantize entire model components (vision encoder, LLM, Q-Former) to $k$-bits.
In addition, we expand our benchmarks to include visual question-answering (VQA) on VQAv2~\cite{goyal2017makingvvqamatter} and GQA~\cite{hudson2019gqanewdatasetrealworld}. Both are successors of prior VQA datasets that contained biases, giving a false sense of understanding.

\subsection{State-of-the-art Quantization Benchmark}
\label{subsec:sota_quant_benchmark}

For each task, we randomly sample a set of 128 image and text pairs from the respective dataset to serve as the calibration set for AWQ and GPTQ. We evaluate BLIP-2 ViT-g for retrieval, and BLIP-2 ViT-g  OPT 2.7B for captioning and VQA tasks. Additionally, we evaluate LLaVA 1.5 7B for VQA tasks. We evaluate on 10\% of the val2014 split for VQAv2, which equates to 21435 samples, and on the entire Test-Dev split for GQA.

It is important to note that typically, when unimodal SOTA quantization methods are applied to MLLMs, only the language model component is quantized. In our search space, we include configurations where we also quantize the vision model and connector components. Our SOTA quantization experiment search space selects bit widths from $\{2,3,4,5,6,8\}$ and considers model components  $C \subseteq \{\text{ViT}, \text{LLM}, \text{Q-Former}\}$ for BLIP-2 and model components $C \subseteq \{\text{ViT}, \text{LLM}\}$ for LLaVA.

We show results for our SOTA experiments in Figure~\ref{fig:sota_plots}. Overall, the unimodal SOTA methods are able to preserve task performance better at more extreme bit widths than uniform quantization. For uniform quantization, the configurations with the lowest bpw that achieved comparable task performance to the full-precision model fell around 6.0-8.0 for retrieval tasks and 8.0-10.0 for captioning. For the SOTA methods, this optimal performance-size trade-off falls into the 3.5-4.5 bpw range for all tasks. We observe that AWQ tends to degrade in model performance more steeply in the extreme bit-width regime than GPTQ for captioning and VQA tasks. This trend can be viewed across both the BLIP-2 and LLaVA architectures. The opposite is true, however, for retrieval where AWQ configurations outperform GPTQ configurations at lower bpw.

\subsection{Component Impact Ablations}
\label{subsec:component_impact_ablations}

\begin{figure}[t]
	\centering
	\includegraphics[width=1.0\linewidth]{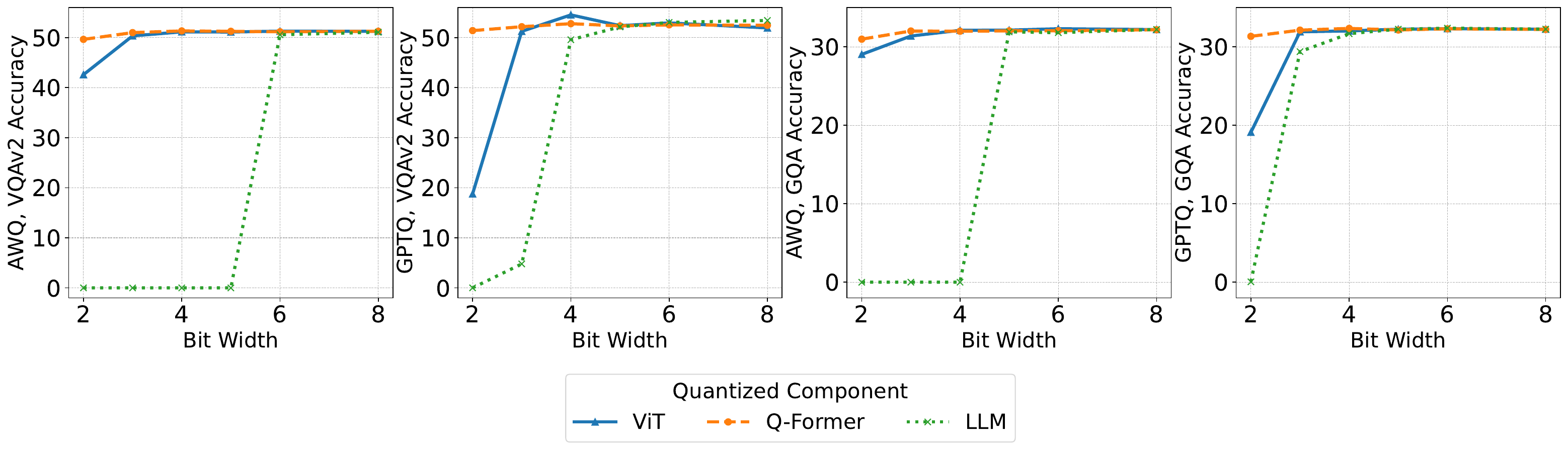}
	\caption{\textbf{Individual quantization} of BLIP-2's ViT, Q-Former, and LLM for VQAv2 and GQA. We show that the different components have varying sensitivities in regards to bit width and quantization method. The LLM tends to have the highest sensitivity in question-answering tasks, followed by the ViT and Q-Former.}
	\label{fig:blip2_vqa_independent}
\end{figure}

\begin{figure}[t]
	\centering
	\includegraphics[width=1.0\linewidth]{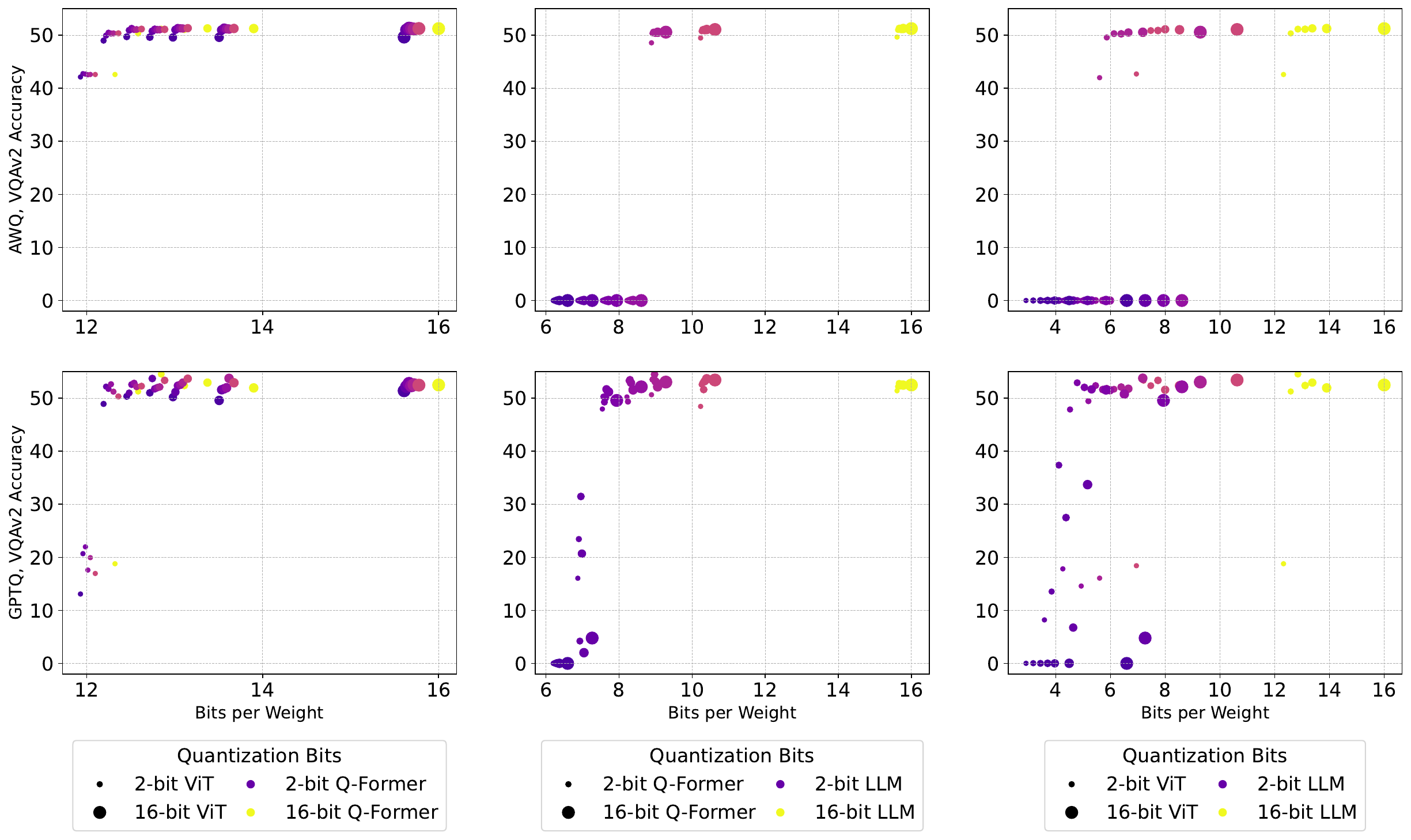}
	\caption{\textbf{Pairwise quantization} of BLIP-2's ViT, Q-Former, and LLM for VQAv2. We draw attention to how components interact with each other when quantized. Notably, the quantization of multiple components yields worse performance than single component quantization.}
	\label{fig:blip2_vqa_pairwise_vqav2}
\end{figure}

From our SOTA quantization experiments, we pulled configurations in which only a single component is quantized for Figure~\ref{fig:blip2_vqa_independent} and configurations in which two components are quantized for Figure~\ref{fig:blip2_vqa_pairwise_vqav2}. Figure~\ref{fig:blip2_vqa_independent} gives an estimate of how each component individually affects the model's performance in VQA. We notice that for both AWQ and GPTQ, the LLM tends to be the most sensitive and drops off earlier than the ViT and the Q-Former. The Q-Former has the least sensitivity, but it also has the fewest parameters. Surprisingly, the ViT experiences a large drop-off with GPTQ, but it experiences relatively little degradation with AWQ.

Figure~\ref{fig:blip2_vqa_pairwise_vqav2} shows the interaction between components when quantized. Empirically, the ViT and LLM tend to have a stronger influence on other components, which could be due to their large parameter sizes. Notably, the LLM has the strongest influence on performance, as seen by the complete loss of performance at around 7 bpw when quantizing the LLM at a lower bit width than the ViT for both VQAv2 and GQA.

\section{Component Importance Analysis}
\label{subsec:component_importance_analysis}

Understanding the relative importance of different components in vision-language models under quantization presents unique analytical challenges. The relationship between bit precision and performance is complex and potentially non-linear, with interactions between components that are difficult to capture with simple models. Our preliminary experiments showed that linear approaches yield poor fits ($R^2 < 0.20$), indicating the need for more sophisticated methods.

\subsection{Setup}
We propose a framework based on three complementary tree-based importance analysis techniques that can effectively capture non-linear relationships and interaction effects. These methods provide different perspectives on component importance, allowing us to establish a consensus ranking that is robust to the limitations of any single approach.

\textbf{Random Forest Feature Importance} requires Random Forest regressors trained to predict performance scores based on the bit precision of each component:

$$
\text{score} = f(\text{vit\_bits}, \text{qformer\_bits}, \text{llm\_bits})
$$

The Random Forest naturally captures non-linear relationships and interactions between features. We extract the built-in feature importance metric, which measures the total reduction in impurity (variance) attributable to each feature across all trees. To quantify uncertainty, we implement bootstrap resampling ($n=100$) to obtain 95\% confidence intervals.

\textbf{Permutation Feature Importance} provides a model-agnostic measure of feature relevance by quantifying how much model performance degrades when a feature's values are randomly shuffled:

$$
I_j = \mathbb{E}_{X,y}[L(f, X^j, y)] - \mathbb{E}_{X,y}[L(f, X, y)]
$$

where $L$ is the performance loss, $X^j$ is the dataset with feature $j$ permuted, and $I_j$ is the importance of feature $j$. By breaking the relationship between the feature and the target, we can measure how much the model relies on that feature for prediction accuracy. We perform 50 permutation iterations and calculate confidence intervals to ensure reliable estimates.

\textbf{SHapley Additive exPlanations (SHAP)} \cite{lundberg2017unifiedapproachinterpretingmodel} provide a game-theoretic approach to feature attribution based on Shapley values. This method calculates the contribution of each feature to each prediction by considering all possible combinations of features:

$$
\phi_j = \sum_{S \subseteq N \setminus \{j\}} \frac{|S|!(|N| - |S| - 1)!}{|N|!} [f_x(S \cup \{j\}) - f_x(S)]
$$

SHAP values have several desirable properties: they sum to the difference between the actual prediction and the average prediction, they are consistent (a model's dependence on a feature can only increase when the feature's importance increases), and they account for feature interactions. We use TreeExplainer \cite{lundberg2019explainableaitreeslocal}, which efficiently computes SHAP values for tree-based models. The mean absolute SHAP value for each component provides a measure of global importance that accounts for both positive and negative effects across the dataset.

\subsubsection*{Consensus Ranking}

To establish a robust, method-agnostic importance ranking, we implement a consensus approach that normalizes and aggregates results across all three feature attribution techniques. For each method (Random Forest, Permutation, SHAP), we first normalize component importance scores to sum to 100\%, ensuring comparable scales. We then average these normalized values across methods to compute a consensus importance percentage for each component (ViT, Q-Former, LLM). This approach mitigates method-specific biases while capturing complementary aspects of feature importance—predictive power from Random Forest, direct performance impact from Permutation analysis, and instance-level contributions from SHAP. The resulting consensus percentages, presented in the appendix, directly inform optimal bit-width allocation strategies across model components.

\subsection{Results}

\begin{figure}[t]
	\centering
	\includegraphics[width=\linewidth]{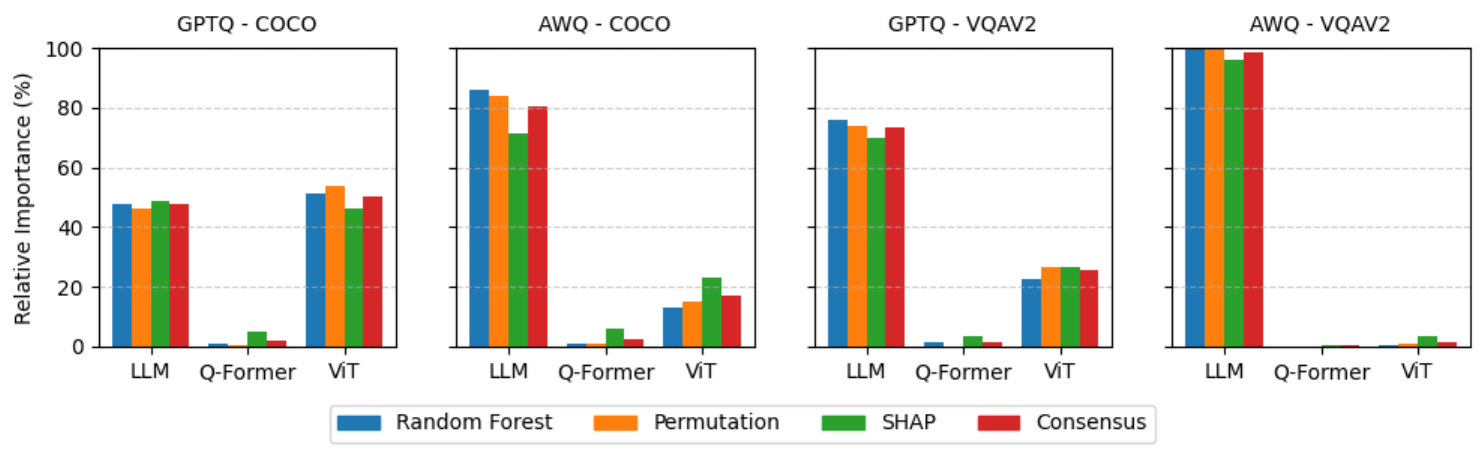}
	\caption{\textbf{BLIP-2 Quantization Component Performance} on COCO and VQAv2 tasks. We report normalized importance percentages of the Vision Transformer (ViT), Q-Former and LLM components under GPTQ and AWQ. Component importance varies across quantization technique and task. GPTQ has a more balanced distribution of importances while AWQ has a stronger skew towards the LLM.}
	\label{fig:feature_importance_blip2}
\end{figure}

\begin{figure}[t]
	\centering
	\includegraphics[width=\linewidth]{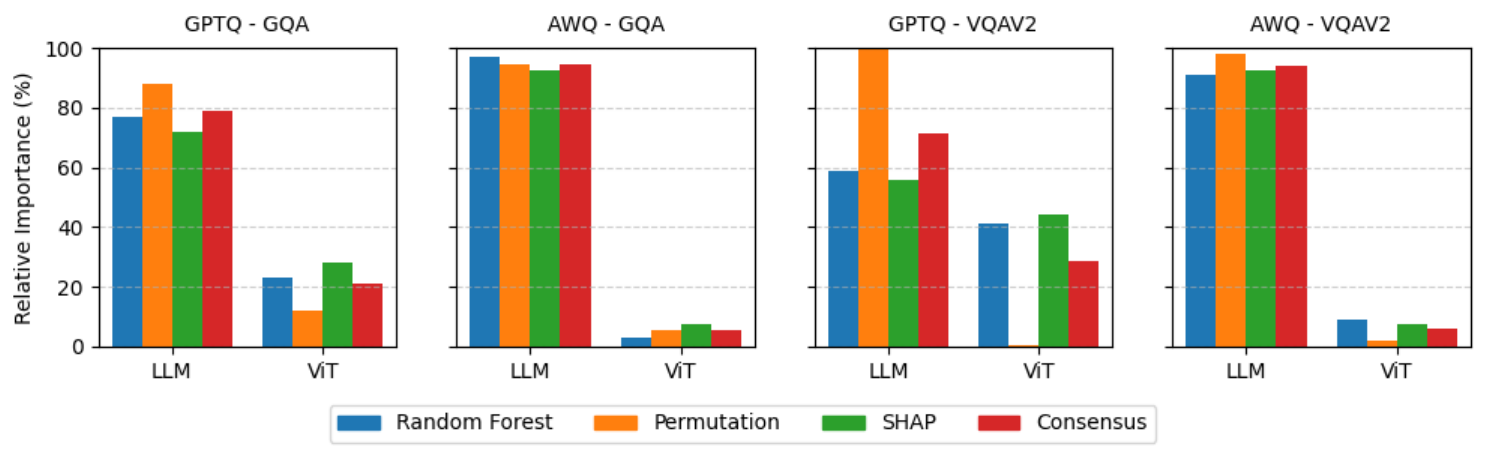}
	\caption{\textbf{LLaVA Quantization Component Performance} on GQA and VQAv2 tasks. We report normalized importance percentages of the model components under GPTQ and AWQ.}
	\label{fig:feature_importance_llava}
\end{figure}

\noindent\textbf{Quantization method dramatically shifts component importance.}
The quantization method significantly affects component importance, with AWQ consistently assigning greater importance to the LLM compared to GPTQ, which distributes effects more evenly across components. In BLIP-2, for COCO captioning, GPTQ balances importance at 50.4\% for the ViT and 47.6\% for the LLM, while AWQ shifts this to 17.1\% for the ViT and 80.5\% for the LLM, as shown in Figure~\ref{fig:feature_importance_blip2}. This difference persists across tasks: GPTQ results in 22-78\% importance for the ViT (depending on dataset), while AWQ results in $\ge73\%$ to the LLM in all cases that utilize it. 

This trend extends to LLaVA as well, where AWQ results in the LLM having a much greater importance after quantization. As shown in Figure~\ref{fig:feature_importance_llava}, AWQ heavily favors the LLM with a 94.62\% and 93.89\% importance to the LLM for both GQA and VQAv2, respectively, compared to GPTQ's more balanced distribution of 79.13\% and 72.15\% LLM importance for GQA and VQAv2, respectively. Despite the ViT accounting for only 4.3\% of the total parameters, it maintains 20-30\% importance under GPTQ for both tasks, proving that importance is not directly proportional to component size.

These shifts suggest AWQ’s activation-aware approach prioritizes the LLM’s large weight matrices, concentrating quantization effects there, while GPTQ’s Hessian-based method captures broader component interactions. Thus, \emph{bit-allocation strategies must adapt to the quantization method}, as AWQ’s LLM focus contrasts with GPTQ’s balanced distribution.

\noindent\textbf{Task characteristics drive component importance variations.}
Task demands strongly dictate component importance, with consensus values varying across datasets. For BLIP-2, the COCO captioning task results in a balanced importance distribution under GPTQ (50.4\% ViT, 47.6\% LLM), suggesting equivalent reliance on both visual encoding and text generation. Retrieval tasks (Flickr), which utilize only the ViT and Q-Former without an LLM component, show ViT dominance under both methods (ViT $\ge70\%$). Notably, the Q-Former's importance increases substantially in these retrieval tasks (15-30\% versus $<3\%$ in other tasks), highlighting its critical role in aligning visual and textual embeddings when no decoder is present to compensate for misalignment. 

Conversely, reasoning-intensive VQA tasks shift importance dramatically toward the LLM. For BLIP-2, the LLM accounts for 73.1-76.7\% importance with GPTQ and 98.3-98.6\% with AWQ on VQAv2 and GQA tasks. In LLaVA, we observe a similar pattern: for VQAv2, GPTQ allocates 72.15\% importance to the LLM and 27.85\% to the vision model, while for GQA, the distribution is 79.13\% LLM and 20.87\% vision model. Under AWQ, this skew becomes even more pronounced, with the LLM accounting for a 94.62\% and 93.89\% importance and the vision model only 5.38\% and 6.11\% for GQA and VQAv2, respectively. These patterns correspond to underlying task requirements: the LLM is primarily responsible for the sophisticated language generation these tasks require.

\noindent\textbf{Architectural layout and component interplay shape quantization patterns.}

The architectural layout and interplay of components strongly shape quantization patterns, redistributing importance based on their roles and sequential dependencies. For BLIP-2’s retrieval tasks (Flickr–Text and Flickr–Image), which exclude the LLM, the Q-Former’s importance jumps to 21.6–29.7\% under GPTQ and 15.3–26.8\% under AWQ, compared to $<3\%$ in tasks with the LLM ( Figure~\ref{fig:feature_importance_blip2}). This shift may indicate the Q-Former’s greater alignment role when no LLM compensates for visual-textual mismatches. In contrast, the Vision Model in LLaVA, connected directly to the LLM via a smaller linear projection layer, shows heightened importance of 20.87–27.85\% under GPTQ for GQA and VQAv2, respectively (Figure~\ref{fig:feature_importance_llava}), possibly because the limited capacity of the projection layer increases the Vision Model’s role in visual processing, despite its mere 4.3\% parameter share.

Pairwise experiments further highlight this interplay: in BLIP-2, quantizing both ViT and LLM simultaneously worsens performance more than individual quantization (Figure~\ref{fig:blip2_vqa_pairwise_vqav2}), revealing non-additive effects from sequential dependencies. The Q-Former’s end-position in retrieval tasks increases its sensitivity, while LLaVA’s Vision-LLM linkage amplifies the Vision Model’s impact. Thus, \emph{bit-allocation strategies must account for architectural context and component dependencies}, beyond method-specific or task-driven shifts, to optimize quantization across diverse VLMs.

\section{Conclusion}

In this work, we systematically investigated the effects of quantization on vision-language models, with a focus on understanding how different components of multimodal architectures like BLIP-2 and LLaVA respond to reduced precision. Our experiments with uniform quantization and state-of-the-art methods such as GPTQ and AWQ reveal that model components exhibit distinct sensitivities to quantization, often with the language model being the most critical, especially in tasks requiring complex reasoning. We demonstrate that SOTA quantization techniques can maintain high performance at significantly lower bit widths compared to uniform approaches, and that the choice of quantization method can substantially alter the relative importance of different model components.

These insights provide practical guidance for optimizing the performance-efficiency tradeoff in multimodal systems, enabling more practical deployment scenarios. 
While our study is limited to simulated quantization without capturing end-to-end latency or hardware-specific optimizations, these constraints point to natural extensions in future work. 
Our open-source contributions of calibration implementations, comprehensive ablation studies, and component analysis tools enable the community to build upon these findings.
As the field expands beyond vision-language to incorporate other modalities, the approach developed in this paper provides a systematic methodology for quantifying component importance across heterogeneous architectures. 
This understanding can directly inform practical tradeoffs in multimodal model compression for resource-constrained environments, from mobile devices to edge computing platforms.

{\small

}


\clearpage
\appendix

\section{Appendix}
\subsection{Uniform Quantization Derivation}
\label{app:uniform_quant}
We define $k$-bit uniform quantization in depth as follows.
Given full-precision weights, $x$, we first normalize them to $[0,1]$ by

$$s(x) = \frac{x - w_{\text{min}}}{w_{\text{max}} - w_{\text{min}}}$$

where $w_{max}$ and $w_{min}$ refer to the per-tensor maximum and minimum weight values of $x$, respectively. 
The normalized weights are then assigned to their closest $k$-bit integer values, yielding discretized weights $\hat{x}$ with

$$ \hat{x} = \frac{1}{2^k -1} \cdot \text{round}((2^k -1) * s(x))$$

Finally, the discretized weights, $\hat{x}$, are returned to their original scale, yielding quantized weights $Q(x)$ with

$$Q(x) = (w_{\text{min}} - w_{\text{max}}) \cdot \hat{x} + w_{\text{min}} $$

\subsection{Optimizing Performance Supplemental}

\begin{figure}[h]
	\centering
	\includegraphics[width=\linewidth]{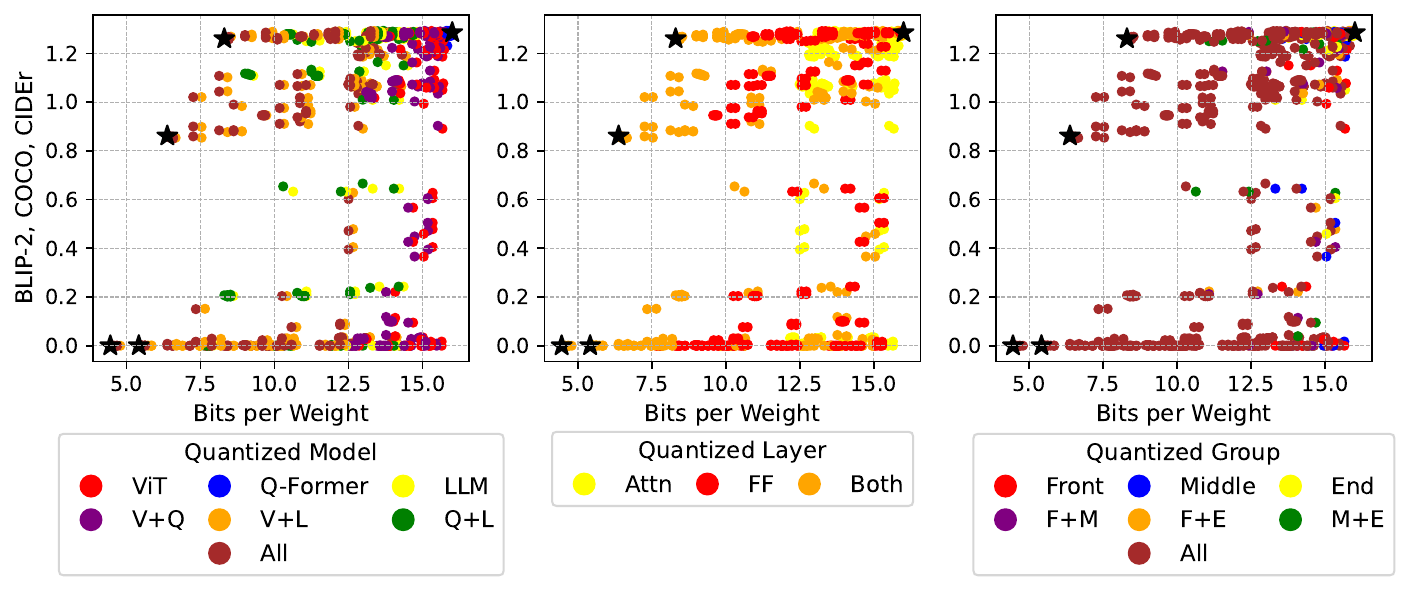}
	\caption{\textbf{Uniform quantization} impact on COCO captioning for BLIP-2.
    We provide full results to complement the zoomed-in results in Figure~\ref{fig:blip_uniform_trio}.
    Black stars show results when we quantize the entire pipeline at 4, 5, 6, 8, and 16 bits.
    The best performance-size tradeoffs tend to come when we quantize the entire pipeline.
    }
	\label{fig:blip_uniform_trio_full_sup}
\end{figure}

\begin{figure}[t]
	\centering
	\includegraphics[width=\linewidth]{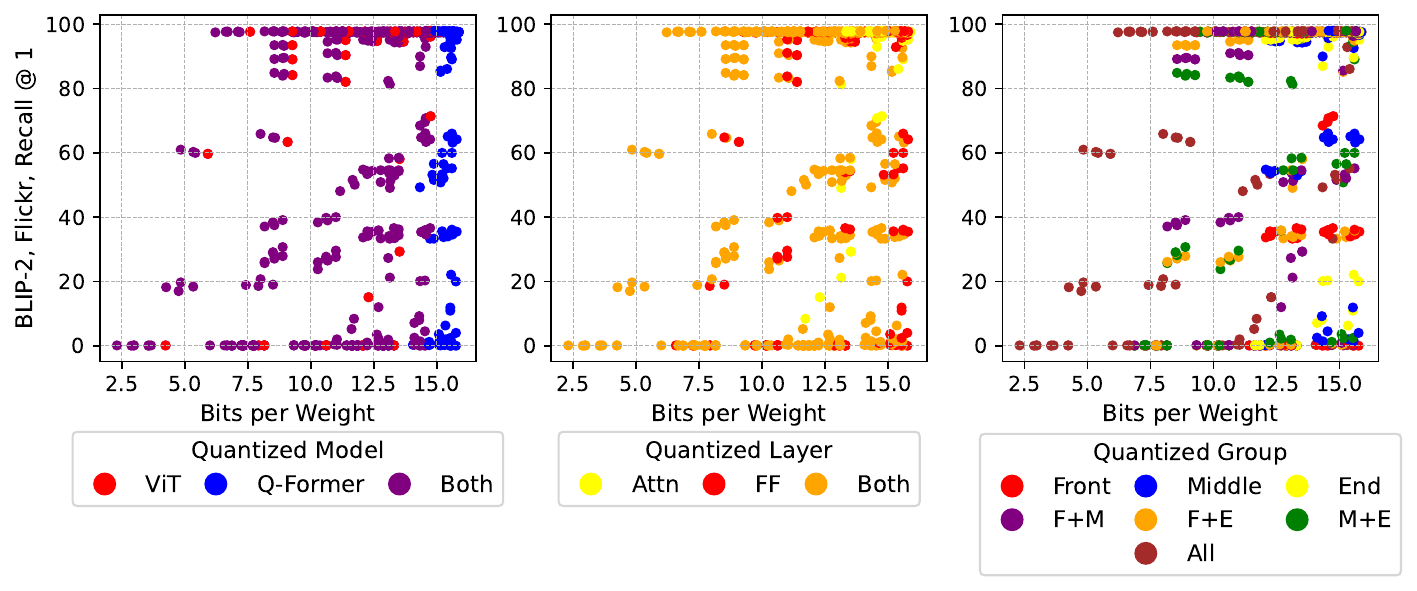}
	\caption{\textbf{Uniform quantization} impact on FLICKR text-to-image retrieval for BLIP-2.
    We draw attention first to which model component we quantize (left), then layer type (middle), then group (right).
    The best performance-size tradeoffs tend to come when we quantize the entire pipeline.
    }
	\label{fig:blip_uniform_trio_flickr_sup}
\end{figure}

We provide Figure~\ref{fig:blip_uniform_trio_full_sup} to complement the results in Figure~\ref{fig:blip_uniform_trio}, this time by showing the full range of bit widths and performances than we test, rather than only focusing on the portion that yields high performance.
We also give results on another task, retrieval in Figure~\ref{fig:blip_uniform_trio_flickr_sup}.
Our findings remain consistent even for this other task.

\begin{figure}[ht]
	\centering
	\includegraphics[width=1.0\linewidth]{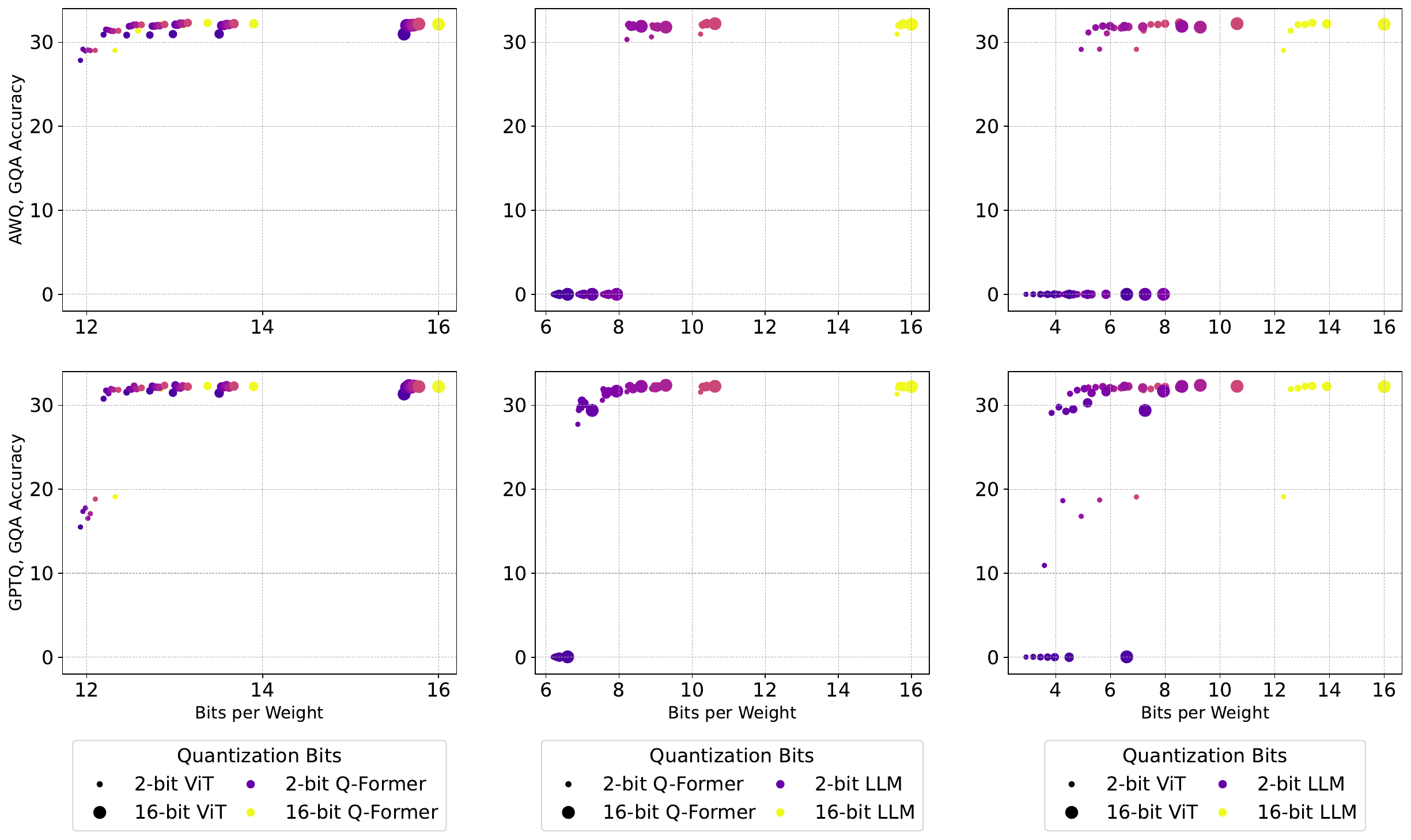}
	\caption{\textbf{AWQ and GPTQ pairwise quantization} of BLIP-2's vision encoder, Q-Former, and LLM for GQA.}
	\label{fig:blip2_vqa_pairwise_gqa}
\end{figure}

We also complement the results in Figure~\ref{fig:blip2_vqa_pairwise_vqav2} with Figure~\ref{fig:blip2_vqa_pairwise_gqa}.
This shows that our findings hold on the GQA dataset.

\subsection{Component Importance Supplemental}

\begin{figure}[t]
	\centering
	\includegraphics[width=\linewidth]{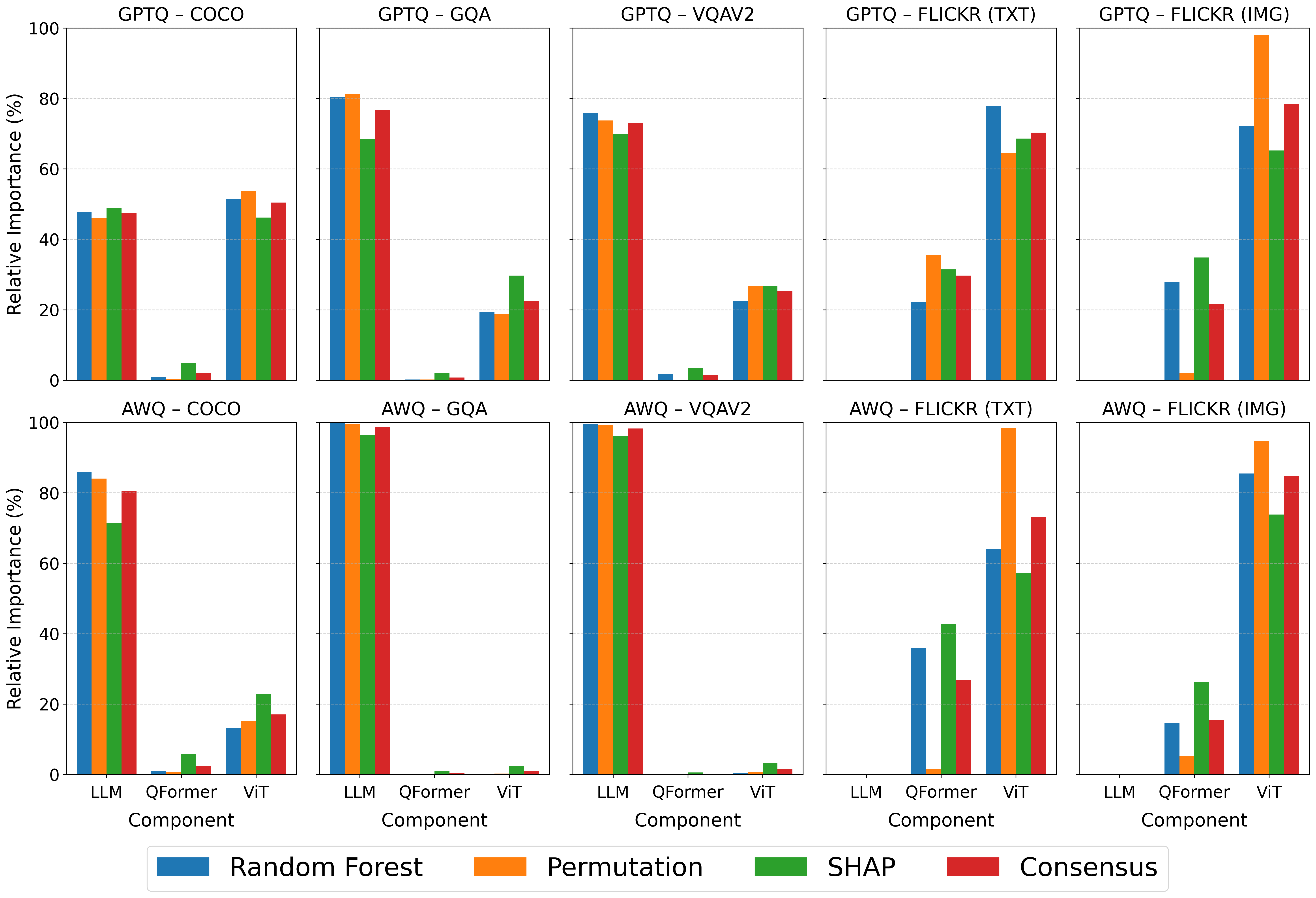}
	\caption{\textbf{Component importance analysis across quantization methods and datasets.} This figure shows the relative importance of different BLIP-2 components (LLM, Vision Transformer, and QFormer) when quantized using GPTQ (top row) and AWQ (bottom row) across five evaluation datasets. The three colors represent different feature importance analysis methods: Random Forest, Permutation, SHAP, and the Consensus between the three.}
	\label{fig:feature_importance_grid}
\end{figure}

We give component importance analysis across multiple tasks and datasets for BLIP-2 in Figure~\ref{fig:feature_importance_grid}.
Our findings remain consistent with those in Figure~\ref{fig:feature_importance_blip2} even considering these additional datasets.
We show this figure mainly for thoroughness, along with Table~\ref{tab:blip2_consensus_importance}, which provides the numbers reflected in these plots.

\begin{table}[ht]
  \centering
  \caption{Consensus feature-importance (\%) of model components.}
  \label{tab:blip2_consensus_importance}
  \begin{tabular}{lllrrr}
    \toprule
    Model & Method & Dataset       & ViT   & QFormer & LLM   \\
    \midrule
    \multirow{10}{*}{BLIP-2} & \multirow{5}{*}{GPTQ} & COCO           & 50.4  &  2.0    & 47.6  \\
                             &                       & GQA            & 22.6  &  0.7    & 76.7  \\
                             &                       & VQAv2          & 25.3  &  1.5    & 73.1  \\
                             &                       & Flickr--Text   & 70.3  & 29.7    & --    \\
                             &                       & Flickr--Image  & 78.4  & 21.6    & --    \\
    \cmidrule{2-6}
                             & \multirow{5}{*}{AWQ}  & COCO           & 17.1  &  2.5    & 80.5  \\
                             &                       & GQA            &  1.0  &  0.4    & 98.6  \\
                             &                       & VQAv2          &  1.5  &  0.2    & 98.3  \\
                             &                       & Flickr--Text   & 73.2  & 26.8    & --    \\
                             &                       & Flickr--Image  & 84.7  & 15.3    & --    \\
    \cmidrule{1-6}
    \multirow{4}{*}{LLaVA} & \multirow{2}{*}{GPTQ}   & GQA            & 20.87  &  --   & 79.13  \\
                           &                         & VQAv2          & 27.85  &  --   & 72.15  \\
    \cmidrule{2-6}
                             & \multirow{2}{*}{AWQ} & GQA             &  5.38  &  --   & 94.62  \\
                             &                      & VQAv2           &  6.11  &  --   & 93.89  \\
    \bottomrule

  \end{tabular}
\end{table}

\end{document}